\documentclass{article}
\usepackage{ijcai20}

\usepackage{times}

\usepackage{soul}
\usepackage{url}
\usepackage[hidelinks]{hyperref}
\usepackage[utf8]{inputenc}
\usepackage[small]{caption}
\usepackage{graphicx}
\usepackage{amsmath}
\usepackage{booktabs}
\usepackage{hyperref}
\usepackage{amsmath,amssymb}

\usepackage{algorithm}
\usepackage{nccmath}
\usepackage{algorithmic}
\usepackage{bm}

\newcommand*\samethanks[1][\value{footnote}]{\footnotemark[#1]}

\title{MAPS: Multi-agent Reinforcement Learning-based Portfolio Management System}

\author{
Jinho Lee\thanks{Equal contribution.}\and
Raehyun Kim\samethanks\and
Seok-Won Yi\And
Jaewoo Kang\footnote{Corresponding author.}\\
\affiliations
Department of Computer Science and Engineering, Korea University
\emails
\{jinholee, raehyun, seanswyi, kangj\}@korea.ac.kr
}

\begin{document}

\maketitle
\thispagestyle{empty}

\begin{abstract}
\textbf{}

Generating an investment strategy using advanced deep learning methods in stock markets has recently been a topic of interest. Most existing deep learning methods focus on proposing an optimal model or network architecture by maximizing return. However, these models often fail to consider and adapt to the continuously changing market conditions. In this paper, we propose the Multi-Agent reinforcement learning-based Portfolio management System (MAPS). MAPS is a cooperative system in which each agent is an independent "investor" creating its own portfolio. In the training procedure, each agent is guided to act as diversely as possible while maximizing its own return with a carefully designed loss function. As a result, MAPS as a system ends up with a diversified portfolio. Experiment results with 12 years of US market data show that MAPS outperforms most of the baselines in terms of Sharpe ratio. Furthermore, our results show that adding more agents to our system would allow us to get a higher Sharpe ratio by lowering risk with a more diversified portfolio.
\end{abstract}

\section{Introduction}\label{sec:intro}
 Most trading decisions nowadays are made by algorithmic trading systems. According to the Deutsche Bank report, the share of automated high-frequency trading in the equity market resulted in a total of 50\% in the US \cite{DeutBank_HFT}.

Decision-making processes based on data analysis are called quantitative trading strategies. Quantitative trading strategies can be divided into two categories: fundamental \cite{Abarbanell1997} and technical analysis \cite{Lo2000,Park2007}. Fundamental analysis refers to performing analysis based on real-world activity. Therefore, fundamental data analysis is mostly based on financial statements and balance sheets. On the other hand, technical analysis is solely based on technical signals, such as historical price and volume. Technicians believe that profitable patterns can be discovered by analyzing historical movements of prices. Traditional quantitative traders attempt to find profitable strategies by constructing algorithms that best represent their beliefs of the market. Although they provide rational clues and theoretical justification of their logic, traditional quantitative strategies are only able to reflect a part of the entire market dynamics. For instance, the momentum strategy \cite{Momentum} assumes that if there exist clear trends, prices will maintain their direction of movement. The mean reversion strategy \cite{MeanRevert} believes that asset prices tend to revert to the average over time. However, it is nontrivial to maintain stable profits under evolving market conditions by leveraging only specific aspects of the financial market. 

Inspired by the recent success of deep learning (DL), researchers have put much effort into finding new profitable patterns from several factors. Early approaches using DL in financial applications focused on how to improve the prediction of stock movements. \citeauthor{DARNN} proposed hierarchical attention combined with a recurrent neural network (RNN) architecture to improve time series prediction. Besides using traditional signals such as stock chart information, there have been numerous attempts to find profitable patterns from new factors like news and sentiment analysis \cite{stocknet}, \cite{event-driven}. More recently, \cite{AdvStock} and \cite{HATS} attempted to create more robust predictions by incorporating adversarial training and corporate relation information, respectively. 

Forecasting models, such as the ones mentioned above, require explicit supervision in the form of labels. These labels take on various forms depending on the task at hand (e.g. up-down-stationary signals for classification). Despite its simple facade, the defining and design of these labels is nontrivial.

Reinforcement learning (RL) approaches provide us with a more seamless framework for decision making \cite{Idiosyncrasies}. The advantage of using RL to make trading decisions is that an agent is trained to maximize its long term reward without supervision.  \citeauthor{FFDR} applied RL with fuzzy learning and recurrent RL. \citeauthor{PracticalRL} proved RL's effectiveness in asset management.

Although the aforementioned work shows promising results, there still remain many challenges in applying DL to portfolio management. Most existing methods utilizing DL focus on proposing a model which simply maximizes expected return without considering risk factors. However, the ultimate goal of portfolio management is to maximize expected return \textit{constrained to a given risk level}, as stated in modern portfolio theory \cite{MPT}. In other words, we must consider risk-adjusted return (e.g. Sharpe ratio) rather than expected return. There has been relatively few work that has considered risk-adjusted return metrics.

In this paper, we propose a cooperative Multi-Agent reinforcement learning-based Portfolio management System (MAPS) inspired by portfolio diversification strategies used in large investment companies. We focus on the fact that investment firms not only diversify assets composing the portfolios, but also the portfolios themselves. Likewise, rather than creating a single optimal strategy, MAPS creates diversified portfolios by distributing assets to each agent. 

Each agent in MAPS creates its own portfolio based on the current state of the market. We designed MAPS' loss function to guide our agents to act as diversely as possible while maximizing their own returns. Agents in MAPS can be seen as a group of independent "investors" cooperating to create a diversified portfolio. With multiple agents, MAPS as a system would have a \textit{portfolio of portfolios}.

We believe that no single strategy fits every market condition, so it is integral to diversify our strategies to mitigate risk and achieve higher risk-adjusted returns. Each agent works towards optimizing a portfolio while keeping in mind that the system as a whole would suffer from a lower risk-adjusted return if they were to create  portfolios similar to that of other agents. Our contribution can be summarized as follows:

\begin{itemize}
	\item To the best of our knowledge, this is the first attempt to use cooperative multi-agent reinforcement learning (MARL) in the field of portfolio management. Given raw financial trading data as the state description, our agents maximize risk-adjusted return.
	\item We devise a new loss function with a diversification penalty term to effectively encourage agents to act as diversely as possible while maximizing their own return. Our experimental results show that the diversification penalty effectively guide our agents to act diversely when creating portfolios.
	\item We conduct extensive experiments on 12 year's worth of US market data with approximately 3,000 companies. The results show that MAPS effectively improves risk-adjusted returns and the diversification of portfolios. Furthermore, we conduct an ablation study and show that adding more agents to our system results in better Sharpe ratios due to further diversification.
\end{itemize}

\section{Problem Statement}\label{sec:prob_form}
In this section, we first introduce the concept of a Markov decision process (MDP) and define how trading decisions are made in a single-agent case. We then extend the single-agent case into a multi-agent case.

\begin{figure}\label{fig:maps-framework}
    \begin{center}
    \includegraphics[width=8cm,height=3.5cm]{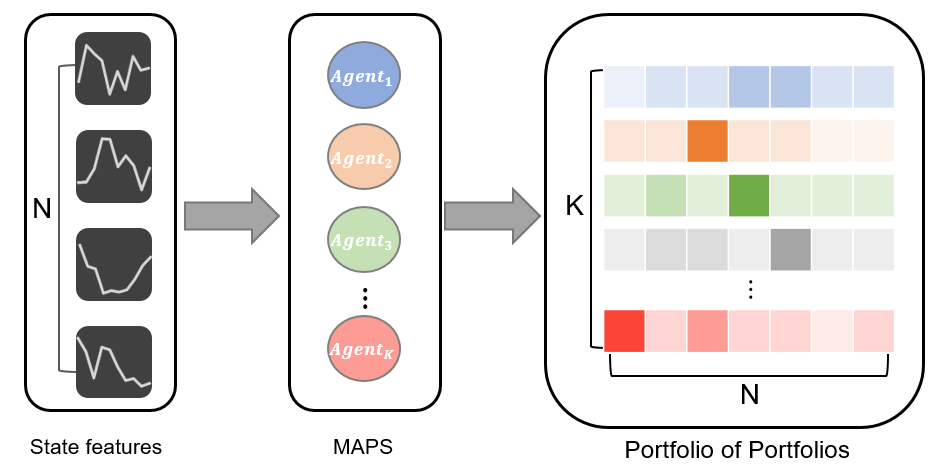}
    \caption{General framework of MAPS.}
    \end{center}
    \vspace{-2mm}
\end{figure}

\subsection{ Single-Agent Reinforcement Learning} 
Single-agent decision-making problems are usually formulated as MDPs. An MDP is defined as a tuple $<s, a, r>$, where $s$ is a finite set of current states, $a$ is a finite set of actions, and $r$ is a reward. The state transition function is omitted for simplicity, since the state transition is not affected by the agent actions in our work. Considering the stochastic and dynamic nature of the financial market, we model stock trading as an MDP as follows:

\begin{itemize}
	\item State $s$: a set of features that describes the current state of a stock. In general, different types of information such as historical price movement, trading volume, financial statements, and sentiment scores can be used as the current state. We use a sequence of closing prices of the past $f$ days of a particular company.
	\item Action $a$: a set of actions. Our agents can take a long, short, or neutral position. 
	\item Reward $r(s,a)$: a reward based on an agent's action at a current state. In this study, a reward is calculated based on the current action and the next day return of a company.
	\item Policy $ \pi(s,a)$: the trading strategy of an agent. A policy $\pi$ is essentially a probability distribution over actions given a state $s$. The goal of an agent is to find the optimal policy which yields maximum cumulative rewards.
\end{itemize}

\subsection{Multi-Agent Reinforcement Learning Extension}
The extension of an MDP to the multi-agent case is called a  stochastic game which is defined as a tuple $<\emph{s}, \textbf{\emph{a}}, \textbf{\emph{r}}>$. Where $s$ is a finite set of current states and \textbf{\emph{a}} is a joint action set  $\textbf{\emph{a}}$ = $a_1$ $\times$ ... $\times$ $a_k$ of $K$ agents. The rewards \textbf{\emph{r}} = $\{$ $r_1$, ...,  $r_k$ $\}$ also depend on current state $s$ and joint action \textbf{\emph{a}} of all agents.
Like the single-agent case, the state transition function is omitted in the multi-agent case. 

In the fully cooperative MARL, the goal of the agents is to find the optimal joint policy $\bm{\pi}$ (s, \textbf{\emph{a}}) to maximize the cumulative rewards \textbf{\emph{r}} of all agents. However, there are two fundamental issues in MARL: the \textit{curse of dimensionality} and the \textit{non-stationarity problem}.

With each additional agent, the joint action space exponentially grows. For example, if one agent can take three total actions (i.e. Long, Neutral, and Short), having ten agents would lead to a total of $3^{10}$ actions. As a result, it becomes more and more difficult to find the optimal joint policy in MARL as the number of agents increases.

In addition, portfolios are comprised of various companies, and an action is typically taken for each company. Considering the combination of all actions for all companies causes the corresponding action space to become exponentially large.

Furthermore, in MARL it is also difficult to find the optimal policy of an agent because all agents learn in conjunction, and consequently the optimal policy of an agent changes as the policy of the other agents change.

Therefore, addressing the proper way to handle the curse of dimensionality problem and designing an appropriate reward structure are central problems in MARL. In the \hyperref[sec:methods]{next section}, we introduce how we handle these problems and consequently how we can effectively guide the agents to act differently from one another while maximizing their own returns.

\begin{figure}\label{fig:maps-network}
    \begin{center}
    \includegraphics[width=8cm,height=3.5cm]{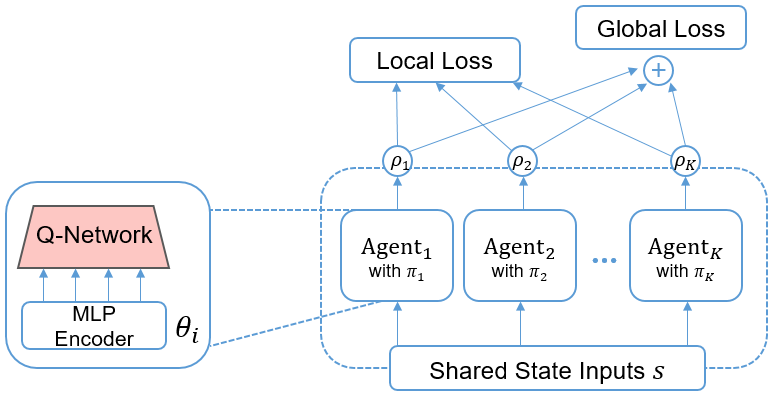}
    \caption{Agent networks and reward strcutrue of MAPS.}
    \end{center}
    \vspace{-2mm}
\end{figure}
\section{Methods}\label{sec:methods}
\subsection{MAPS Architecture}
In this section, we describe the overall architecture of MAPS which is illustrated in \hyperref[fig:maps-network]{Figure 2}. In MAPS, all of our $K$ agents are trained via Deep Q-learning \cite{DQN}. Each agent consists of an MLP encoder and Q-network, with structures varying from agent to agent. The input to each agent is a shared state $s$ which is a vector of length $f$. Each $s$ consists of the normalized closing price sequences of the past $f$ days. The output vector of each agent, $\rho$, is a vector of length 3 with each element representing the expected long term reward of actions Long, Neutral, and Short, respectively, given the current state $s$. Therefore, an MLP encoder maps raw state features provided from the environment (i.e. the closing price sequence of a company) into an action value.

To handle the curse of dimensionality and the non-stationarity problem mentioned in the previous section, we use following methods. First, when calculating the action values of agent $i$ the other agents' actions are ignored. Doing so limits the possible number of total actions to three (i.e. Long, Neutral, and Short). Second, each agent maintains two MLP network parameter sets, $\theta$ and $\theta^*$. The network parameter set $\theta$ is used when performing the gradient step to minimize loss, and the target network parameter set $\theta^*$ is simply a copy of $\theta$, and is updated periodically to handle the non-stationarity problem due to the changes of policies of other agents during training. We also adopted experience replay \cite{DQN} to reduce correlation between subsequent episodes.

The overall training procedure is as follows. For each iteration, the episode for each agent is sampled using an  $\epsilon$-greedy policy \cite{suttonRL} from a training data set of size $N$ $\times$ $T$, stored in a memory buffer of size $K$ $\times$ $M$, where $N$, $T$, and $M$ each indicates the number of companies, the number of days, and the size of the memory buffer of each agent. Then, a batch of size $K$ $\times$ $\beta$ is sampled from the memory buffer to calculate the loss. Finally, the gradient step is performed to minimize loss with respect to the parameters $\theta$. $\theta$ is copied to $\theta^*$ after every $C$ iterations ($C \in \mathbb{Z}$).

\subsection{Shared State Memory Buffer}

The first step in our training procedure is to sample an episode $e_i^m$ from the training data and to store it in the memory buffer. Unlike the single-agent case, the memory buffer is a $K$ $\times$ $M$ matrix where $K$ is the number of agents. 

An episode $e_i^m$ is a tuple defined as $e_i^m$ = $<$ $s_c^t$, $a_{i,c}^t$, $r_{i,c}^t$, $s_c^{t+1}$ $>$, where $i$ and $m$ denote the index of an agent and the column index of the memory buffer. $s_c^t$ and  $a_{i,c}^t$ each refer to the current state of company $c$ at time $t$ and the action chosen by the $\epsilon$-greedy policy of agent $i$ given current state $s_c^t$, and $r_{i, c}^t$ and $s^{t + 1}_c$ each refer to the immediate reward received by agent $i$ and the subsequent state of company $c$ at time $t + 1$. Note that there is no subscript index for the agents in $s_c^t$ and $s_c^{t+1}$. This means that the same input state is stored in the same column in the memory buffer.

An action $a_{i,c}^t$ and reward $r_{i,c}^t$ are defined as follows.    

\begin{equation} \label{eq:action}\tag{1}
    a_{i,c}^t = 1 - \operatorname*{argmax} \{\rho_{i,c}^t\}
\end{equation}

\begin{equation} \label{eq:reward}\tag{2}
    r_{i,c}^t = a_{i,c}^t\times R_c^t
\end{equation}
where $\rho_{i,c}^t$ and $R_c^t$ refer to the output vector of agent $i$ given input $s_c^t$ and the daily return of company $c$ between time $t$ and time $t+1$ represented in percentage. Therefore, the value 1, 0, or -1 is assigned to action $a_{i,c}^t$ for Long, Neutral, or Short actions, respectively.

The next step is to sample random batches of size $K \times \bm{\beta}$ from the memory buffer to calculate loss. To formulate the procedure, we define a sampled batch as a $K \times \bm{\beta}$ matrix and define a vector $r$ of length $\bm{\beta}$. At every iteration, a random integer value in the range $[0, \bm{\beta})$ is sampled and assigned to vector $r$. Then the element at the $i$th row and $b$th column in the batch matrix is assigned as: $h_{i}^b$ $\leftarrow$ $e_{i}^{r_b}$, where ${r_b}$ indicates $b$th element in vector $r$. 

The intuition behind this sampling method is to share the same input state sequence among agents in each batch. Since vector $r$ is re-sampled every iteration rather than by each agent, the same column index sequence is sampled from the memory buffer for each agent. Consequently, as shown in \hyperref[fig:maps-network]{Figure 2}, every agent is trained using the same input sequence and we can therefore guide the agents to act different from each other despite being given identical input sequences.

\subsection{Loss Function}

As previously mentioned, our goal is to guide the agents in MAPS to act as diversely as possible while maximizing their own rewards. To achieve these two contradicting goals, we design our loss function to have two components, namely a \textit{local loss}, and a \textit{global loss}. The \textit{local loss} of each agent is calculated based only on the reward and action value of a particular agent. We first define $LLoss_i^b$ of agent $i$ calculated using a single episode at the $i$th row and $b$th column in the batch matrix.

\begin{equation} \label{eq:reward_loss}\tag{3}
LLoss_i^b = \left[ Q(s_i, a_i; \theta_i) - r +  \gamma \underset{a_i'}{\max} Q(s_i', a_i'; \theta_i^*)\right]^2
\end{equation} 
where $s_i$, $a_i$, $r_i$, $s_i'$, and $a_i'$ each indicate the current state, current action, immediate reward, next state, and next action, respectively, and $Q$ refers to the action-value function. These values are obtained from episode $h_{i}^b$ in the batch matrix. Note that while choosing action $a_i'$ given state $s_i'$, the target network of agent $i$ parameterized by $\theta_i^*$ is used to avoid the moving target problem. We get \textit{local loss} by summing up the $LLoss_i^b$ over batch size $\bm{\beta}$ as follows:

\begin{equation} \label{eq:local_loss}\tag{4}
    LLoss_i = \sum_{b=1}^{\beta}  LLoss_i^b
\end{equation}

However, it is not possible an agent to be aware of the actions of other agents with the local reward alone. Therefore, the \textit{global loss} provides additional guidance to our agents. We define the \textit{positional confidence} score of agent $i$ for company $c$ calculated using a single episode at the $i$th row and $b$th column of the batch matrix as follows:

\begin{equation} \label{eq:pos_conf_score}\tag{5}
    \eta_{i,c}^b = \rho_{i,c}[\text{Long}] - \rho_{i,c}[\text{Short}]
\end{equation}
where $\rho_{i,c}$ is the output vector of agent $i$ given input $s_c$. Since the elements of $\rho_{i,c}$ each represent the actions of agent $i$, respectively, $\eta_{i,c}^b$ represents the $i$th agent's confidence of how much company $c$'s  price will rise at the subsequent time step. By concatenating the calculated positional confidence scores, we get a positional confidence vector of agent $i$:

\begin{equation}\tag{5}
\eta_i = \left[\eta_i^1, ..., \eta_i^{\beta}\right]^T
\end{equation}

We penalize similar behavior among the agents by minimizing the correlation of positional confidence vectors between agents. Formally, the \textit{global loss} can be expressed as:
\begin{equation} \label{eq:group_loss}\tag{6}
    GLoss_i = \sum_{i=1, i\neq j}^K \left[\text{Corr}(\eta_i, \eta_j^*)\right]^2
\end{equation}

Note that while creating a positional confidence vector of agent $j$ for a agent $i$, we use the target network parameterized by $\theta_j^*$ to mitigate the effect of the non-stationarity problem.

Finally, our total loss is a weighted sum of the \textit{local loss} and the \textit{global loss}. 

\begin{equation} \label{eq:comb-loss}\tag{7}
    Loss_i = (1-\lambda)LLoss_i + \lambda GLoss_i\quad 
\end{equation}
where $\lambda$ is a hyperparameter with a value within $[0, 1]$. The training procedure is summarized in \textbf{Algorithm 1}. The value of \emph{maxiter}, $\bm{\beta}$ and $C$ are 400,000, 128, and 1000, respectively.

\begin{algorithm}
\caption{ Training algorithm }
\begin{algorithmic}[1]

\FOR{ \emph{maxiter} }
    \STATE Store experience $e_i^m$ in the memory buffer. \\
    \STATE Sample batch of size $K$ $\times$ $\bm{\beta}$ from the memory buffer. \\
    \STATE Calculate $Loss_i$ for each agent. \\
    \STATE Perform gradient descent to minimize $Loss_i$ w.r.t. the parameters $\theta_i$ for each agent. \\
    \STATE Copy $\theta_i^*$ $\leftarrow$ $\theta_i$ at every $C$ iterations for each agent.
\ENDFOR
\end{algorithmic}
\end{algorithm}

\subsection{Portfolio of Portfolios}
When training is finished, each of our agents is expected to output an action value. We create a final portfolio vector $\bm{\alpha^t}$ at time $t$ by summing the portfolio vectors of each agent. The portfolio vector of agent $i$ at time $t$ (i.e. $\alpha^t_i$) is a vector of length $N$, which satisfies $ \sum_{c=1}^N |\alpha^t_{i,c}| = 1$, where $\alpha^t_{i,c}$ represents the $c$th element in the vector  $\alpha^t_i$. Thus, each $\alpha^t_{i,c}$ represents the weight assigned to company $c$ at time $t$ by agent $i$. We use the positional confidence score $\eta_{i,c}$ to create the portfolio vector of agent $i$ at time $t$ as follows:

\begin{equation} \label{eq:normalize}\tag{8}
    \alpha^t_{i,c} = \dfrac{\eta^t_{i,c}}{\sum_{c=1}^N |\eta^t_{i,c}|}
\end{equation}

Note that superscript $t$ is added to $\eta_{i, c}$ (i.e. $\eta^t_{i, c}$) since the test is proceeded on a test set size of $N$ $\times$ $T$, not on the batch. The final portfolio vector $\bm{\alpha^t}$ is calculated as follows.

\begin{equation} \label{eq:normalize_port}\tag{9}
    \bm{\alpha^t_{c}} = \frac{\sum_{i=1}^K \alpha^t_{i,c}} {K}
\end{equation}
where $\bm{\alpha^t_c}$ represents the $c$th element in vector $\bm{\alpha^t}$. Finally, the portfolio vector $\bm{\alpha^t}$ is normalized to satisfy $ \sum_{c=1}^N |\bm{\alpha^t_c}|$ = 1.0.

\section{Experiments}\label{sec:exp}
\subsection{Experimental Settings}
\begin{table}[]\label{table:dataset}
\centering
\begin{tabular}{|c|c|c|c|c|}
\hline
      & Period    & N    & \#Data  \\ \hline
Training & 2000-2004 & 1534 & 1876082 \\ \hline
Validation & 2004-2006 & 1651 & 779272  \\ \hline
Test  & 2006-2018 & 2061 & 6019248 \\ \hline
\end{tabular}
\caption{The statistics of dataset}
\end{table}
\paragraph{\textbf{Dataset}} We collected roughly 18 year's worth of daily closing price data of approximately  3,000 US companies. Specifically, we used the list of companies from the  Russell 3000 index. 

We divided our dataset into training set validation set and test set. Detailed statistics of our dataset are summarized in \hyperref[table:dataset]{Table 1.} The validation set is used to optimize the hyperparameters. 

\paragraph{\textbf{States \& Hyperparameters}} Among many possible candidates, we gave our agents raw historical closing prices as state description features. However, it is worth noting that our framework is not restricted to certain types of state features, and other kinds of features such as technical indicators or sentiment scores can also be used. We expect further diversification to occur if various sources of information were to be provided to MAPS, and leave this as an open question for future work. 

\textbf{MAPS@\textit{k}} is our proposed model with \textit{k} agents in the system. \textit{k} is an arbitrary hyperparameter and we choose among the values [4, 8, 16] for our experiments to show the effect of using different numbers of agents.

To explain the structure of the MLP encoder, we take \textbf{MAPS@4} as an example. An MLP of size $[32, 16]$ represents agent \#1, and each subsequent agent has an extra layer with double the hidden units. For example, agent \#2 would be an MLP of size $[64, 32, 16]$. \textbf{MAPS@8} and \textbf{MAPS@16} are simply structures where this pattern is repeated two and four times, respectively. 

Batch normalization \cite{batch} is used after every layer except the final layer and Adam optimizer \cite{adam} was used with a learning rate of 0.00001 to train our models. The value of $\lambda$ was empirically chosen as $0.8$ based on the validation set. 

\paragraph{\textbf{Evaluation Metric}} We measure profitability of methods with Return and Sharpe ratio. 

\begin{itemize}
    \item \textbf{Return} We calculated the daily return of our portfolio $\bm{\alpha^t}$ as follows:
    \begin{equation*} \label{eq:return}\tag{5.1}
              R^{\alpha}_{t} = 100 \times \sum_{c=1}^N\left( \frac{p_c^{t+1} - p_c^t}{ p_c^t}\right) \cdot \bm{\alpha_c^{t}}
    \end{equation*}
    where $p_t^c$ denotes the closing price of stock $c$ at time $t$. 

\end{itemize}

\begin{itemize}
    \item \textbf{Sharpe Ratio} The annualized Sharpe ratio is used to measure the performance of an investment compared to its risk. The ratio calculates the excess earned return to the risk-free rate per unit of volatility (risk) as follows:
    \begin{equation*} \label{eq:sharpe}\tag{5.2}
                 \text{Sharpe} =  \sqrt{252} \times \frac{\mathrm{E}[R^{\alpha}_{t} - R^f_t]}{\text{std}[R^{\alpha}_{t}-R^f_t]}
    \end{equation*}  
    where $R^f_t$ is daily risk-free rate at time $t$ and 252 is the number of business days in a year.
    
\end{itemize}

\paragraph{\textbf{Baselines}} We compare MAPS with the Russell 3000, which is one of the major indices, and the following baselines:

\begin{itemize}
	\item \textbf{Momentum (MOM)} is an investment strategy based on the belief that current market trends will continue. We use the simplest version of the strategy: the last 10-day price movements are used as momentum indicators.
	
	\item \textbf{Mean-Reversion (MR)} strategy works on the assumption that there is a stable underlying trend line and the price of an asset changes randomly around this line. MR believes that asset prices will eventually revert to the long-term mean. The 30-day moving average is used as the mean reversion indicator.
	
    \item \textbf{MLP, CNN} Among many existing stock movement forecasting methods, we chose these two models as our forecast baselines as they are widely used in stock forecasting \cite{Forecast-survey}. The MLP model in our experiments consists of five hidden layers with sizes of [256, 128, 64, 32, 16]. The CNN model has four convolutional layers with [16, 16, 32, 32] filters and one fully connected layer of size [32] is used. Max-pooling layers are applied after the second and fourth layers. Batch normalization is applied for both models. Both models have one additional prediction layer with a softmax function and are trained with 3-label (i.e. up, neutral, down) cross-entropy loss.
	
	\item \textbf{DA-RNN} refers to the dual-stage attention-based RNN \cite{DARNN}. Is is the state-of-the-art and attention mechanisms are used in each stage to identify relevant input features and select relevant encoder hidden states. As the DA-RNN model was originally designed to forecast time series signals, we also trained our model to predict future prices of the assets with mean squared error. A portfolio is created based on the expected return of the predicted asset prices.

\end{itemize}

\subsection{Results}

\begin{table}[]
\centering
\begin{tabular}{@{}|c||cc||cc|@{}}
\toprule
Period       & \multicolumn{2}{c||}{2006-2012}    & \multicolumn{2}{c|}{2012-2018}  \\ \hline\hline
Models        & Return            & Sharpe             & Return            & Sharpe             \\ \hline\hline
MOM      & \phantom{-}3.938          & \phantom{-}1.149          & -3.223          & -1.198               \\
MR           & -2.262          & -0.899          & \phantom{-}2.220          & \phantom{-}0.816               \\
MLP           & \phantom{-}16.377          & \phantom{-}1.309          & \phantom{-}1.744          & \phantom{-}0.368               \\
CNN            & \phantom{-}17.036          & \phantom{-}1.093          & -3.294          & -0.442               \\ 
DA-RNN            & \phantom{-}11.860          & \phantom{-}3.283          & \phantom{-}4.309          & \phantom{-}2.113               \\\hline\hline
MAPS@4           & \phantom{-}17.955          & \phantom{-}4.829          & \phantom{-}4.846          & \phantom{-}2.121          \\
MAPS@8       & \phantom{-}22.744          & \phantom{-}4.751          & \phantom{-}\textbf{6.123}          & \phantom{-}2.175          \\ 
MAPS@16 & \phantom{-}\textbf{23.467} & \phantom{-}\textbf{5.547} & \phantom{-}5.567 & \phantom{-}\textbf{2.247} \\ 
\bottomrule
\end{tabular}
\caption{Experimental results on the Russell 3000 companies.}
\label{tab:result-rt}
\end{table}

\begin{table}[]
\centering
\begin{tabular}{@{}|c||c||c|@{}}
\toprule
Models       & {2006-2012}    & {2012-2018}  \\ \hline\hline

MAPS@4      & 0.3415 & 0.3456  \\
MAPS@8      & 0.4622 & 0.4424  \\ 
MAPS@16     & 0.2318 & 0.2429 \\ 
\bottomrule
\end{tabular}
\vspace{-3mm}
\caption{Average correlation of daily return of each agents. A smaller correlation indicates more independent actions of agents.}
\label{tab:result-cor}
\end{table}

\paragraph{\textbf{Performance analysis}}
Our experiment results are summarized in \hyperref[tab:result-rt]{Table 2}, and Figure 3 illustrates a comparison of cumulative wealth based on the portfolios created by each model. MAPS outperformed all baselines in terms of both annualized return and Sharpe ratio. Some interesting findings are as follows:

\begin{itemize}
	\item The performance of traditional strategies like MOM and MR vary based on market conditions and generally perform poorly. As these strategies use a single rule leveraging only certain aspects of market dynamics, their performance is not robust as the market evolves.
    \item Forecast-based methods show  better performance than traditional approaches in terms of annualized return. Naturally, the performance of forecast-based methods heavily relies on the prediction accuracy of the model. The MLP and CNN perform better in general but did not always outperform the traditional methods. Only DA-RNN performed consistently better in both annualized return and Sharpe ratio for both testing periods.
    
    \item MAPS outperformed all baselines in our experiments. \textbf{It is worth noting} that MAPS shows a better Sharpe ratio even when the return is similar. This proves the effectiveness of diversification with multiple agents in the perspective of risk-adjusted return. We further observe that MAPS with more agents obtains a better Sharpe ratio. 
    \item One unexpected result is that the Sharpe ratio does not scale linearly with the number of agents. We can interpret this with the daily return correlation scores of different MAPS,  summarized in \hyperref[tab:result-cor]{Table 3}. If our agents act diversely, the correlation of daily return would be small. In \hyperref[tab:result-cor]{table 3} we can find that the average correlation of MAPS@8 is higher than MAPS@4. The agents of MAPS@8 act more similarly to each other than those of MAPS@4, resulting in lower Sharpe ratio despite higher returns. Further improvement in our learning scheme may solve this issue and we leave this for future work. 
\end{itemize}

\begin{figure}\label{fig:cum-rts}
    \begin{center}
    \includegraphics[width=8.6cm,height=4.8cm]{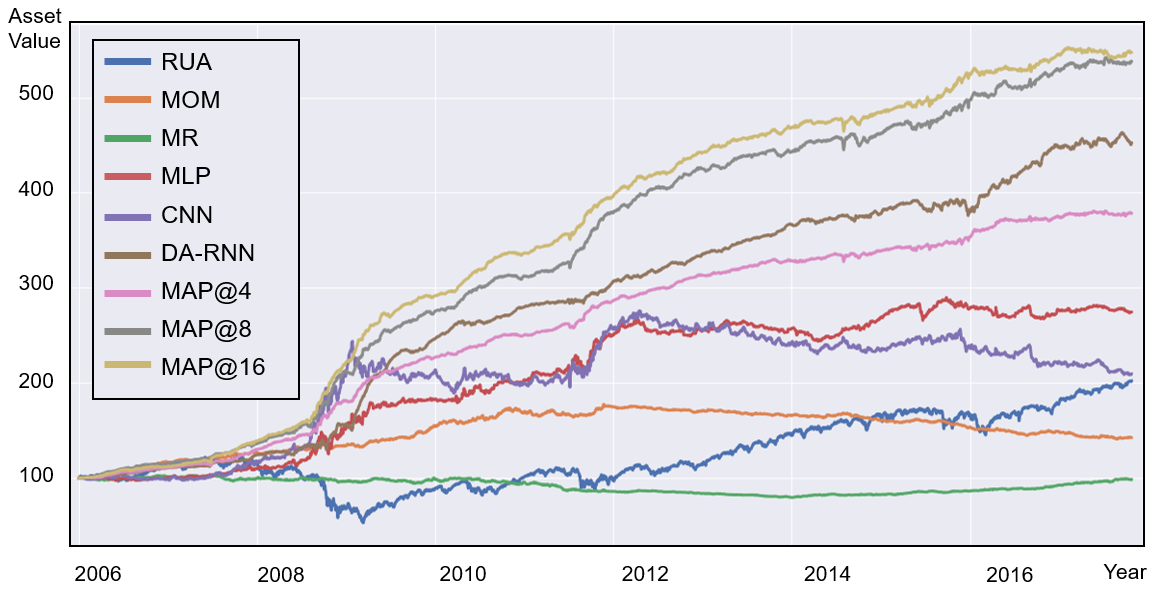}
    \caption{Comparison of cumulative wealth of different models (RUA indicates the Russell 3000 index). All asset values are set to 100 at the beginning of the test period.}
    \end{center}
    \vspace{-2mm}
\end{figure}

\begin{figure}\label{fig:global-loss}
    \begin{center}
    \includegraphics[width=8.6cm,height=4cm]{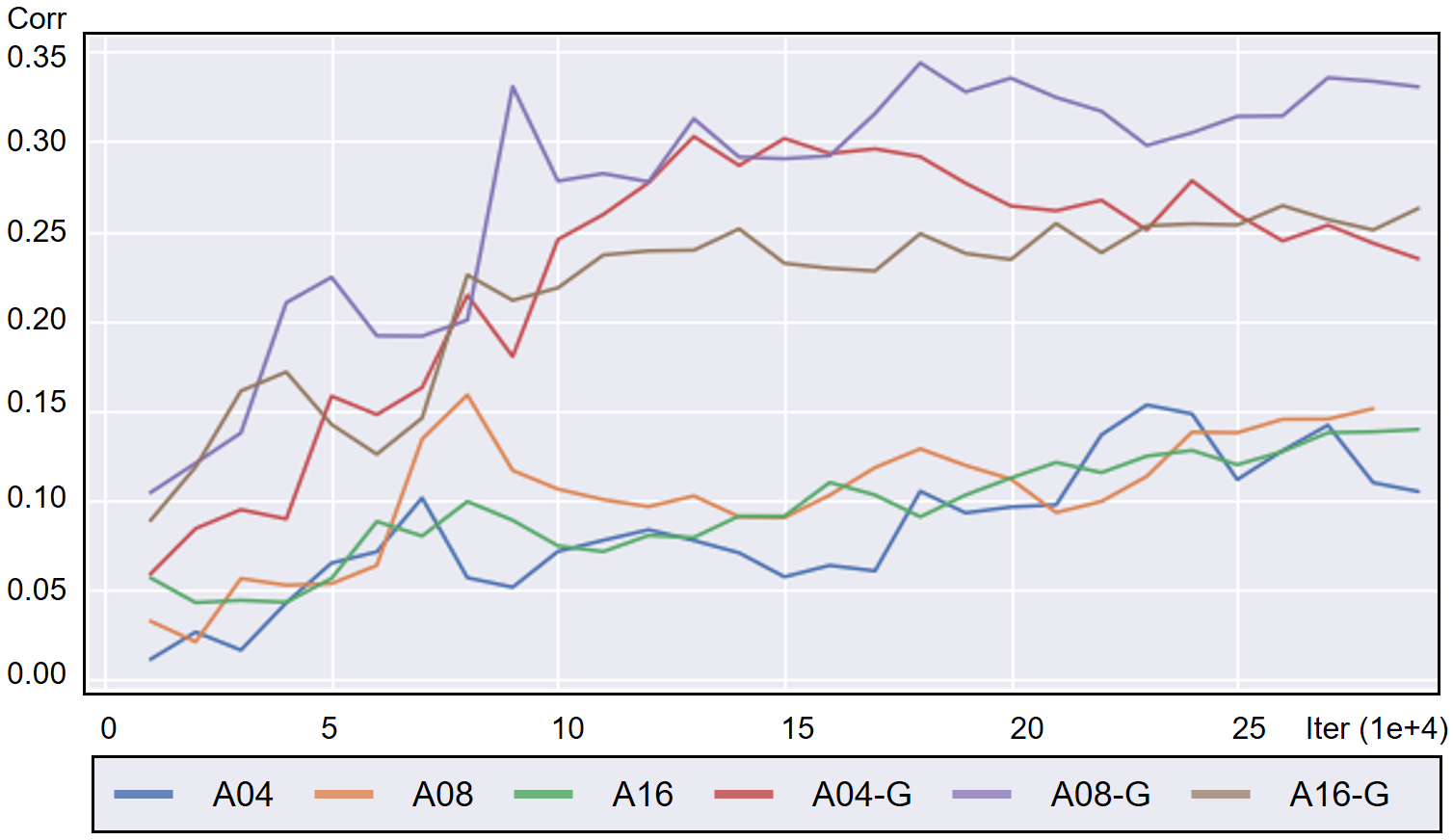}
    \caption{Illustrations of how agents in MAPS choose different actions given same state features.}
    \end{center}
    \vspace{-2mm}
\end{figure}

\paragraph{\textbf{Effect of \textit{global loss}}} To investigate the effect of a \textit{global loss}, we compare the learning process of MAPS with and without \textit{global loss}. During our learning process, we calculate the correlation of a  positional confidence score $\eta_i$ between all agents for the entire validation set. We calculate this value every 10,000 training iterations and average for all companies and pairs of agents. As the positional confidence value indicates the type of action taken by the agents, higher correlation means more similar actions among the agents. As we can see in \hyperref[fig:global-loss]{Figure 4.}, average correlation values of MAPS without \textit{global loss} increase rapidly and converge with much higher values than MAPS with \textit{global loss}. In contrast, the correlations of MAPS trained with \textit{global loss} increased slowly and resulted in having small values. The results verify the effectiveness of \textit{global loss} in making agents act independently.

\begin{figure}\label{fig:agent-port}
    \begin{center}
    \includegraphics[width=8.5cm,height=4.7cm]{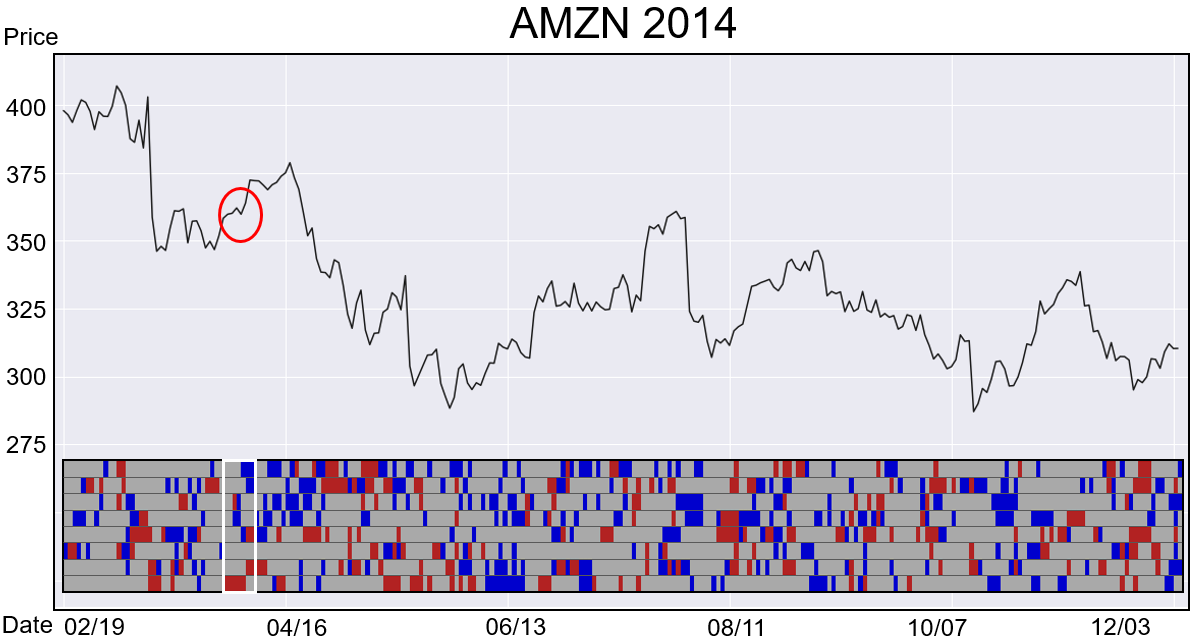}
    \caption{Illustrations of how agents in MAPS choose different actions given same state features.}
    \end{center}
    \vspace{-2mm}
\end{figure}

\paragraph{\textbf{Case Study}} To better understand how our agents act differently with identical state features, we illustrate an example portfolio in \hyperref[fig:agent-port]{Figure 5.} The black line is a movement of Amazon's stock price in 2016. The colored rectangle at the bottom of the figure describes the actions of our agents. In this case, we have eight agents in MAPS and each line represents which positions were taken by which agent, with each color representing a position. Red, grey, and blue each refer to long, neutral, and short positions taken at a given time. What we can observe here is that the agents in our system make different decisions based on their own understanding of the market. For instance, in the spring of 2014 (the period we outlined with the bright white box), the future movement of Amazon stock price seems volatile and uncertain after a steep fall and several price corrections. Two out of eight agents decided to take long positions betting that the future price would rise, and two agents chose a short position with the opposite prospect. This kind of discrepancy in actions is prevalent throughout the trading process, making our portfolio as a whole sufficiently diversified. 
\section{Conclusion}\label{sec:conclusion}
In this work, we propose MAPS, a cooperative Multi-Agent reinforcement learning-based Portfolio management System. The agents in MAPS act as differently as possible while maximizing their own reward guided by our proposed loss function. Experiments with 12 years of US market data show that MAPS outperforms most of the existing baselines in terms of Sharpe ratio. We also presented the effectiveness of our learning scheme and how our agents' independent actions end up with a diversified portfolio with detailed analysis. 

\section*{Acknowledgements}
This work was supported by the National Research Foundation of Korea   (NRF-2017R1A2A1A17069645, NRF-2017M3C4A7065887).
\clearpage
\bibliographystyle{named}
\bibliography{ijcai20}

\begin{thebibliography}{}

\bibitem[\protect\citeauthoryear{Abarbanell and Bushee}{1997}]{Abarbanell1997}
Jeffrey~S Abarbanell and Brian~J Bushee.
\newblock Fundamental analysis, future earnings, and stock prices.
\newblock {\em Journal of Accounting Research}, 35(1):1--24, 1997.

\bibitem[\protect\citeauthoryear{Bacoyannis \bgroup \em et al.\egroup
  }{2018}]{Idiosyncrasies}
Vangelis Bacoyannis, Vacslav Glukhov, Tom Jin, Jonathan Kochems, and Doo~Re
  Song.
\newblock Idiosyncrasies and challenges of data driven learning in electronic
  trading.
\newblock {\em arXiv preprint arXiv:1811.09549}, 2018.

\bibitem[\protect\citeauthoryear{Deng \bgroup \em et al.\egroup }{2016}]{FFDR}
Yue Deng, Feng Bao, Youyong Kong, Zhiquan Ren, and Qionghai Dai.
\newblock Deep direct reinforcement learning for financial signal
  representation and trading.
\newblock {\em IEEE transactions on neural networks and learning systems},
  28(3):653--664, 2016.

\bibitem[\protect\citeauthoryear{Di~Persio and Honchar}{2016}]{Forecast-survey}
Luca Di~Persio and Oleksandr Honchar.
\newblock Artificial neural networks architectures for stock price prediction:
  Comparisons and applications.
\newblock {\em International journal of circuits, systems and signal
  processing}, 10(2016):403--413, 2016.

\bibitem[\protect\citeauthoryear{Ding \bgroup \em et al.\egroup
  }{2015}]{event-driven}
Xiao Ding, Yue Zhang, Ting Liu, and Junwen Duan.
\newblock Deep learning for event-driven stock prediction.
\newblock In {\em Twenty-fourth international joint conference on artificial
  intelligence}, 2015.

\bibitem[\protect\citeauthoryear{Feng \bgroup \em et al.\egroup
  }{2019}]{AdvStock}
Fuli Feng, Huimin Chen, Xiangnan He, Ji~Ding, Maosong Sun, and Tat-Seng Chua.
\newblock Enhancing stock movement prediction with adversarial training.
\newblock In {\em Proceedings of the 28th International Joint Conference on
  Artificial Intelligence}, pages 5843--5849. AAAI Press, 2019.

\bibitem[\protect\citeauthoryear{Ioffe and Szegedy}{2015}]{batch}
Sergey Ioffe and Christian Szegedy.
\newblock Batch normalization: Accelerating deep network training by reducing
  internal covariate shift.
\newblock {\em arXiv preprint arXiv:1502.03167}, 2015.

\bibitem[\protect\citeauthoryear{Jegadeesh and Titman}{1993}]{Momentum}
Narasimhan Jegadeesh and Sheridan Titman.
\newblock Returns to buying winners and selling losers: Implications for stock
  market efficiency.
\newblock {\em The Journal of finance}, 48(1):65--91, 1993.

\bibitem[\protect\citeauthoryear{Kaya \bgroup \em et al.\egroup
  }{2016}]{DeutBank_HFT}
Or{\c{c}}un Kaya, Jan Schildbach, and Deutsche~Bank Ag.
\newblock High-frequency trading.
\newblock {\em Reaching the limits, Automated trader magazine}, 41:23--27,
  2016.

\bibitem[\protect\citeauthoryear{Kim \bgroup \em et al.\egroup }{2019}]{HATS}
Raehyun Kim, Chan~Ho So, Minbyul Jeong, Sanghoon Lee, Jinkyu Kim, and Jaewoo
  Kang.
\newblock Hats: A hierarchical graph attention network for stock movement
  prediction.
\newblock {\em arXiv preprint arXiv:1908.07999}, 2019.

\bibitem[\protect\citeauthoryear{Kingma and Ba}{2014}]{adam}
Diederik~P Kingma and Jimmy Ba.
\newblock Adam: A method for stochastic optimization.
\newblock {\em arXiv preprint arXiv:1412.6980}, 2014.

\bibitem[\protect\citeauthoryear{Lo \bgroup \em et al.\egroup }{2000}]{Lo2000}
Andrew~W Lo, Harry Mamaysky, and Jiang Wang.
\newblock Foundations of technical analysis: Computational algorithms,
  statistical inference, and empirical implementation.
\newblock {\em The journal of Finance}, 55(4):1705--1765, 2000.

\bibitem[\protect\citeauthoryear{Markowitz}{1952}]{MPT}
Harry Markowitz.
\newblock Portfolio selection.
\newblock {\em The journal of finance}, 7(1):77--91, 1952.

\bibitem[\protect\citeauthoryear{Mnih \bgroup \em et al.\egroup }{2015}]{DQN}
Volodymyr Mnih, Koray Kavukcuoglu, David Silver, Andrei~A Rusu, Joel Veness,
  Marc~G Bellemare, Alex Graves, Martin Riedmiller, Andreas~K Fidjeland, Georg
  Ostrovski, et~al.
\newblock Human-level control through deep reinforcement learning.
\newblock {\em Nature}, 518(7540):529, 2015.

\bibitem[\protect\citeauthoryear{Park and Irwin}{2007}]{Park2007}
Cheolho Park and Scott~H Irwin.
\newblock What do we know about the profitability of technical analysis?
\newblock {\em Journal of Economic Surveys}, 21(4):786--826, 2007.

\bibitem[\protect\citeauthoryear{Poterba and Summers}{1988}]{MeanRevert}
James~M Poterba and Lawrence~H Summers.
\newblock Mean reversion in stock prices: Evidence and implications.
\newblock {\em Journal of financial economics}, 22(1):27--59, 1988.

\bibitem[\protect\citeauthoryear{Qin \bgroup \em et al.\egroup }{2017}]{DARNN}
Yao Qin, Dongjin Song, Haifeng Chen, Wei Cheng, Guofei Jiang, and Garrison
  Cottrell.
\newblock A dual-stage attention-based recurrent neural network for time series
  prediction.
\newblock {\em arXiv preprint arXiv:1704.02971}, 2017.

\bibitem[\protect\citeauthoryear{Sutton and Barto}{2018}]{suttonRL}
Richard~S Sutton and Andrew~G Barto.
\newblock {\em Reinforcement learning: An introduction}.
\newblock 2018.

\bibitem[\protect\citeauthoryear{Xiong \bgroup \em et al.\egroup
  }{2018}]{PracticalRL}
Zhuoran Xiong, Xiao-Yang Liu, Shan Zhong, Hongyang Yang, and Anwar Walid.
\newblock Practical deep reinforcement learning approach for stock trading.
\newblock {\em arXiv preprint arXiv:1811.07522}, 2018.

\bibitem[\protect\citeauthoryear{Xu and Cohen}{2018}]{stocknet}
Yumo Xu and Shay~B Cohen.
\newblock Stock movement prediction from tweets and historical prices.
\newblock In {\em Proceedings of the 56th Annual Meeting of the Association for
  Computational Linguistics (Volume 1: Long Papers)}, pages 1970--1979, 2018.

\end{thebibliography}

\end{document}